\documentclass[sigconf]{acmart}

\copyrightyear{2024}
\acmYear{2024}
\setcopyright{acmlicensed}\acmConference[CIKM '24]{Proceedings of the 33rd ACM International Conference on Information and Knowledge Management}{October 21--25, 2024}{Boise, ID, USA}
\acmBooktitle{Proceedings of the 33rd ACM International Conference on Information and Knowledge Management (CIKM '24), October 21--25, 2024, Boise, ID, USA}
\acmDOI{10.1145/3627673.3680074}
\acmISBN{979-8-4007-0436-9/24/10}

\usepackage{multirow, multicol, caption}
\usepackage{adjustbox}
\usepackage[ruled,noend]{algorithm2e} 
\DontPrintSemicolon
\usepackage{algorithmic}
\usepackage{bm}
\usepackage[utf8]{inputenc}
\usepackage[cjk]{kotex}
\usepackage[textsize=small]{todonotes}
\usepackage{subfigure}
\usepackage{booktabs}
\usepackage{lipsum}
\usepackage[switch]{lineno}
\usepackage{bm}
\usepackage{adjustbox}
\usepackage{subfigure}
\usepackage{booktabs}
\usepackage{multirow}
\usepackage{multicol}
\usepackage{graphicx}
\usepackage{lipsum} 
\usepackage[utf8]{inputenc}
\usepackage{kotex}
\usepackage{siunitx}
\usepackage{rotating}
\usepackage{wrapfig}

\AtBeginDocument{%
  }

\begin{document}

\title{Bridging Dynamic Factor Models and Neural Controlled Differential Equations for Nowcasting GDP}

\author{Seonkyu Lim}
\authornote{Both authors contributed equally to this research.}
\email{sklim@kftc.or.kr}
\affiliation{%
  \institution{Korea Financial Telecommunications and Clearings Institute}
  \city{Seoul}
  \country{South Korea}
}

\author{Jeongwhan Choi}
\authornotemark[1]
\email{jeongwhan.choi@yonsei.ac.kr}
\affiliation{%
  \institution{Yonsei University}
  \city{Seoul}
  \country{South Korea}
}

\author{Noseong Park}
\authornote{Corresponding author}
\email{noseong@kaist.ac.kr}
\affiliation{%
  \institution{Korea Advanced Institute of Science and Technology}
  \city{Daejeon}
  \country{South Korea}
}

\author{Sang-Ha Yoon}
\email{syoon@kiep.go.kr}
\affiliation{%
  \institution{Korea Institute for International Economic Policy}
  \city{Sejong}
  \country{South Korea}
}

\author{ShinHyuck Kang}
\email{shinkang@kipf.re.kr}
\affiliation{%
  \institution{Korea Institute of Public Finance}
  \city{Sejong}
  \country{South Korea}
}

\author{Young-Min Kim}
\author{Hyunjoong Kang}
\email{{injesus, kanghj}@etri.re.kr}
\affiliation{%
  \institution{Electronics and Telecommunications Research Institute}
  \city{Daejeon}
  \country{South Korea}
}

\renewcommand{\shortauthors}{Seonkyu Lim et al.}

\begin{abstract}
Gross domestic product (GDP) nowcasting is crucial for policy-making as GDP growth is a key indicator of economic conditions. Dynamic factor models (DFMs) have been widely adopted by government agencies for GDP nowcasting due to their ability to handle irregular or missing macroeconomic indicators and their interpretability. However, DFMs face two main challenges: i) the lack of capturing economic uncertainties such as sudden recessions or booms, and ii) the limitation of capturing irregular dynamics from mixed-frequency data. To address these challenges, we introduce \texttt{NCDENow}, a novel GDP nowcasting framework that integrates neural controlled differential equations (NCDEs) with DFMs. This integration effectively handles the dynamics of irregular time series.\texttt{NCDENow} consists of 3 main modules: i) factor extraction leveraging DFM, ii) dynamic modeling using NCDE, and iii) GDP growth prediction through regression. 
We evaluate \texttt{NCDENow} against 6 baselines on 2 real-world GDP datasets from South Korea and the United Kingdom, demonstrating its enhanced predictive capability. Our empirical results favor our method, highlighting the significant potential of integrating NCDE into nowcasting models. Our code and dataset are available at \url{https://github.com/sklim84/NCDENow_CIKM2024}.
\end{abstract}

\ccsdesc[500]{Computing methodologies~Machine learning approaches}
\ccsdesc[500]{Applied computing~Economics}
\ccsdesc[500]{Information systems~Data mining}

\keywords{GDP nowcasting, dynamic factor model, neural controlled differential equations}

\maketitle

\section{Introduction}

\begin{figure}[t!]
    \centering
    \includegraphics[width=0.75\columnwidth]{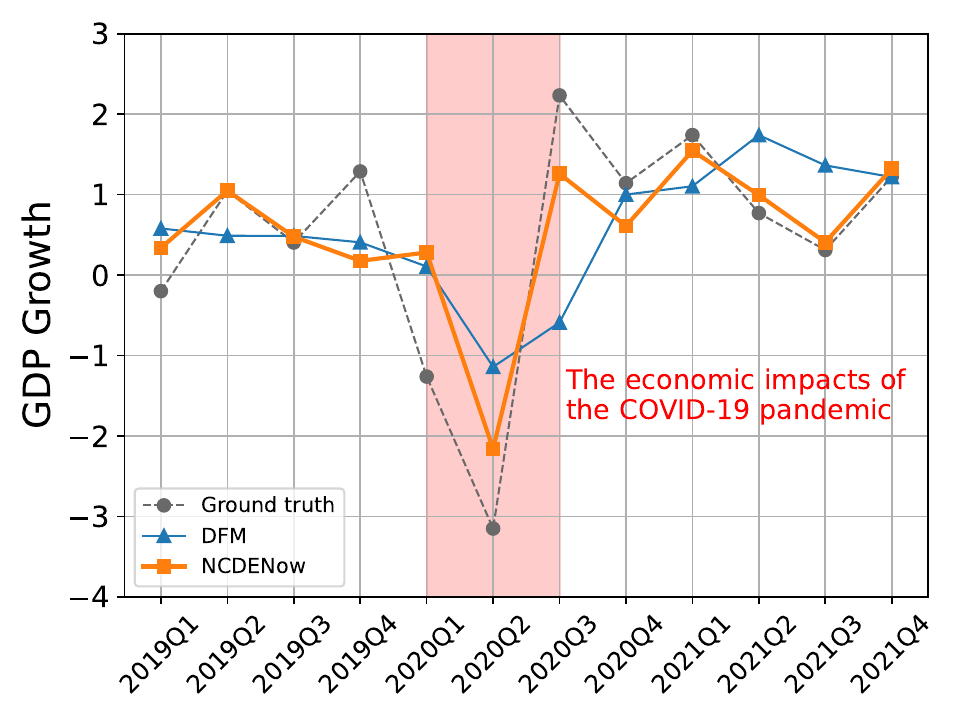}
    \caption{Nowcasting vs. ground-truth on South Korea GDP during COVID-19. \texttt{NCDENow}, our proposed method, captures sudden drops better than DFM.}
    \label{fig:recession}
\end{figure}

Nowcasting GDP growth is a crucial task for policymakers and analysts who need timely information about the current state of the economy to make decisions. However, official GDP growth estimates are released with a significant delay and are subject to revisions. 
In general, the first advance estimates of GDP growth are published about 30 days after the end of the quarter. At the same time, a wealth of information about different aspects of the economy can be obtained from relevant indicators that are released with less or more frequent release delays. For example, governmental institutions such as central banks publish various monthly indicators of economic activity. As a result, the demand for early economic estimates modeled using these more frequent observations has increased, and these initial estimates are called ``nowcasting''.

\begin{figure}[t!]
    \centering
    \includegraphics[width=0.8\columnwidth]{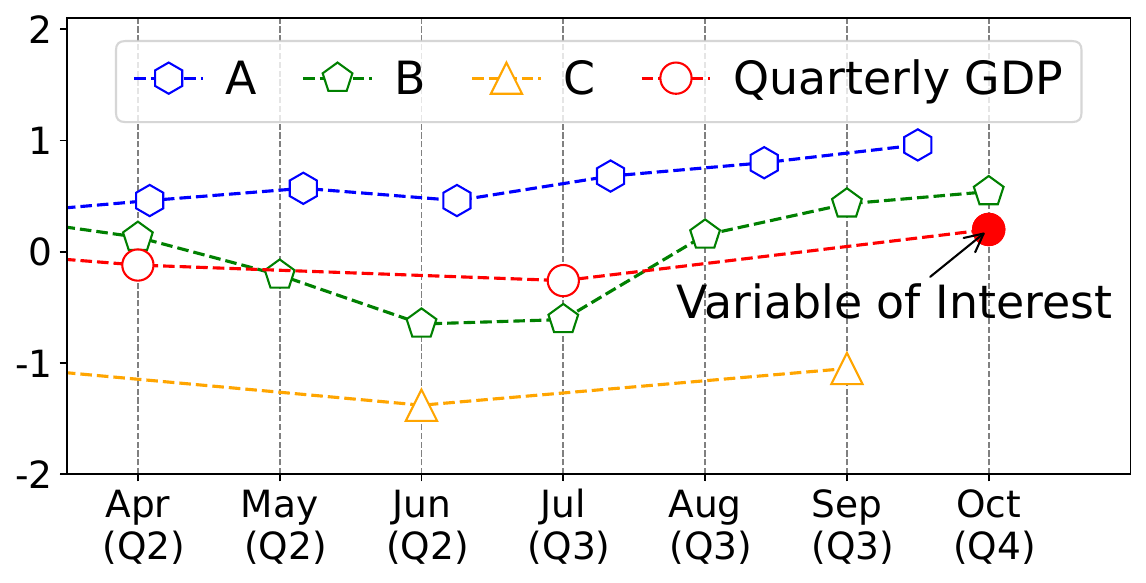}
    \caption{Example of GDP nowcasting. The macroeconomic indicators have multi-frequencies. 
    We want to estimate third-quarter (Q3) GDP growth, but the first estimates are unavailable until the end of October. In this case, we nowcast GDP growth using other indicators released in Q3.}
    \label{fig:example}
\end{figure}

GDP nowcasting is very important for Korean governments because it helps them monitor the impact of their policies, adjust their financial and monetary plans accordingly, and respond quickly to any shocks or crises that may affect the economy. The predicted GDP growth can also inform the public and private sectors about the economic outlook and expectations, which can influence their consumption and investment decisions.
Therefore, economic research on GDP nowcasting~\cite{higgins2014gdpnow,woloszko2020adaptive,lee2017s,lee2022d,wu2021data,nemeth2023gdp,jardet2022nowcasting,assunccao2022nowcasting,cohen2023nowcasting,luo2021determining,ashwin2021nowcasting,banbura2013now,giannone2008nowcasting} has been actively being conducted for ten years and has focused on 3 main issues: 
i) how one incorporates data into a model with missing observations caused by mixed or irregular sampling frequencies, ii) how one handles large numbers of variables, and iii) how one allows for time-varying parameters. 

One of the most prominent methods for nowcasting GDP growth is the use of DFM, which extracts a small number of latent factors representing economic conditions from mixed and irregular time-series data. 
While there has been remarkable progress in using DFM for nowcasting, they face two challenges: i) difficulty in capturing sudden fluctuations and complex patterns (cf. Fig.~\ref{fig:recession}) due to the lack of non-linear activation, and ii) suboptimal handing of missing values from mixed frequency data.
Real-world economic data, even within the same monthly or quarterly indicator, may be released at different times, causing time lags compared to typical sequential data (cf. Fig.~\ref{fig:example}). 
NCDEs, which can theoretically generalize RNNs and state-space models~\cite{cirone2024theoretical}, have recently gained attention as a possible complement to address these limitations.

CDE is one of the most suitable ones for building macroeconomic models. CDE was first proposed by financial mathematicians to model complicated dynamics in financial markets, which is a specific application domain of DFM since financial transactions have a few latent factors. 
NCDE~\cite{kidger2020neural} is a set of techniques to learn CDE from data with neural networks. Note that NCDE keeps reading the time-derivative of data over time, and for this reason, NCDE is, in general, considered as \emph{continuous} RNN. In addition, NCDE is known to be superior in processing irregular time series.

We propose \texttt{NCDENow}, a novel GDP nowcasting framework that integrates DFM with NCDE. The novelty of \texttt{NCDENow} lies in its strategic design, which synergizes the interpretability of DFMs with the temporal modeling capabilities of NCDEs. 
To the best of our knowledge, this is the first research to integrate NCDE with DFMs for economic indicators, offering extensibility to DFM variants.

\texttt{NCDENow} consists of 3 modules: i) the factor extraction module, which extracts latent factors through DFM; ii) the exposure estimation module, which uses NCDE to estimate factor loadings and idiosyncratic disturbances; and iii) the regression module, which integrates the results from these modules to nowcast GDP growth.

We evaluate \texttt{NCDENow} against 6 baselines using GDP datasets from South Korea and the UK. Our method, \texttt{NCDENow}, shows outstanding performance in terms of GDP nowcasting, handling missing data and sudden drop, and interpretability. Our contributions can be summarized as follows:
\begin{itemize}
    \item We propose \texttt{NCDENow}, a novel framework that integrates the strengths of DFM strengths in extracting latent factors with the capabilities of NCDE in handling irregular time series data (Section~\ref{sec:method}).
    \item \texttt{NCDENow} outperforms the baselines in terms of nowcasting accuracy (Section~\ref{subsec:rq1}), particularly in scenarios with irregular data (Section~\ref{subsec:rq2}) and volatile economic conditions such as COVID-19 (Section~\ref{subsec:rq3}).
    \item \texttt{NCDENow} provides meaningful insights into the relationships between extracted factors and GDP growth (Section~\ref{subsec:rq4}), thereby contributing to a deeper understanding of economic conditions and assisting policymakers in making informed decisions.
\end{itemize}

\section{Preliminaries \& Related Work}
This section presents recent GDP nowcasting research and provides background for understanding.

\subsection{GDP Nowcasting}
The Atlanta Fed introduces GDPNow~\cite{higgins2014gdpnow}, which combines the econometric approach from top-down GDP nowcasting with a detailed bottom-up approach. 
The New York Fed's online GDP nowcasting\footnote{https://www.newyorkfed.org/research/policy/nowcast\#/nowcast} uses DFM and a big data framework~\cite{baker2023reintroducing,almuzara2023new,blazqueznowcasting}.
Similarly, \citet{hayashi2023nowcasting} also employ DFM for GDP nowcasting.
The OECD integrates insights from macroeconomics to construct improved tree-based models~\cite{woloszko2020adaptive}.
\citet{wu2021data} employ LSTM with comprehensive datasets.
However, these models face challenges under high economic uncertainty, such as COVID-19.

Recently, the Bank of Korea introduces a GDP nowcasting system~\cite{lee2022d} that combines DFM and LSTM. They use DFM only for data imputation and rely on LSTM for nowcasting, which may not effectively capture the dynamics of macroeconomic indicators.

\subsection{Dynamic Factor Models}

DFM extends the factor model for cross-sectional data to a time-series domain~\cite{geweke1977dynamic}. It assumes that a small number of latent dynamic factors drive the co-movement of high-dimensional time-series data, influenced by idiosyncratic disturbances with zero mean. These factors are estimated by assuming an autoregressive (AR) process~\cite{stock2011dynamic}.
The equation of DFM is as follows:
\begin{align}\label{eq:obs_eq}
    \mathbf{y}_t &= \mathbf{\Lambda} \mathbf{z}_t + \bm{\epsilon}_t, \\
    \mathbf{z}_t &= \mathbf{A}_1\mathbf{z}_{t-1} + \dots + \mathbf{A}_p\mathbf{z}_{t-p} + \bm{\mu}_t, \label{eq:tran_eq}
\end{align} where $\mathbf{y}_t$ is observed data at time $t$, and $\bm{\epsilon}_t$ is idiosyncratic disturbance at time $t$. $\mathbf{z}_t$ is latent dynamic factor at time $t$, and $\bm{\mu}_t$ is factor disturbance at time $t$. $\mathbf{\Lambda}$ is matrix of factor loadings, and $\mathbf{A}_i$ is matrix of autoregression coefficient. Eq.~\eqref{eq:obs_eq} is referred to as the observation equation, while Eq.~\eqref{eq:tran_eq} is denoted as the transition equation.

The representation of a few factors is an effective approach to capturing large-scale economic indicators. Therefore, DFM is one of the primary tools for macroeconomists. \citet{banbura2010nowcasting} propose dividing observed data into groups based on domain knowledge and extracting factors for each group.
~\citet{banbura2010nowcasting, mariano2010coincident} demonstrate the capability to handle mixed frequency data.
However, DFM's assumption of linear relationships between variables limits its ability to capture nonlinearities present in real-world data, particularly during high economic uncertainty.

Recently, various models have been proposed to enhance DFM by leveraging machine learning or deep learning methods. \citet{bontempi2017dynamic} combine DFM principles with deep learning for nonlinear modeling and multi-step-ahead forecasting.
\citet{duan2022factorvae} propose FactorVAE that probabilistically extracts factors based on the variational autoencoder (VAE).
While these studies aim to improve forecasting, we focus on nowcasting GDP growth.

\begin{figure*}[t]
    \centering
    \includegraphics[width=0.8\textwidth]{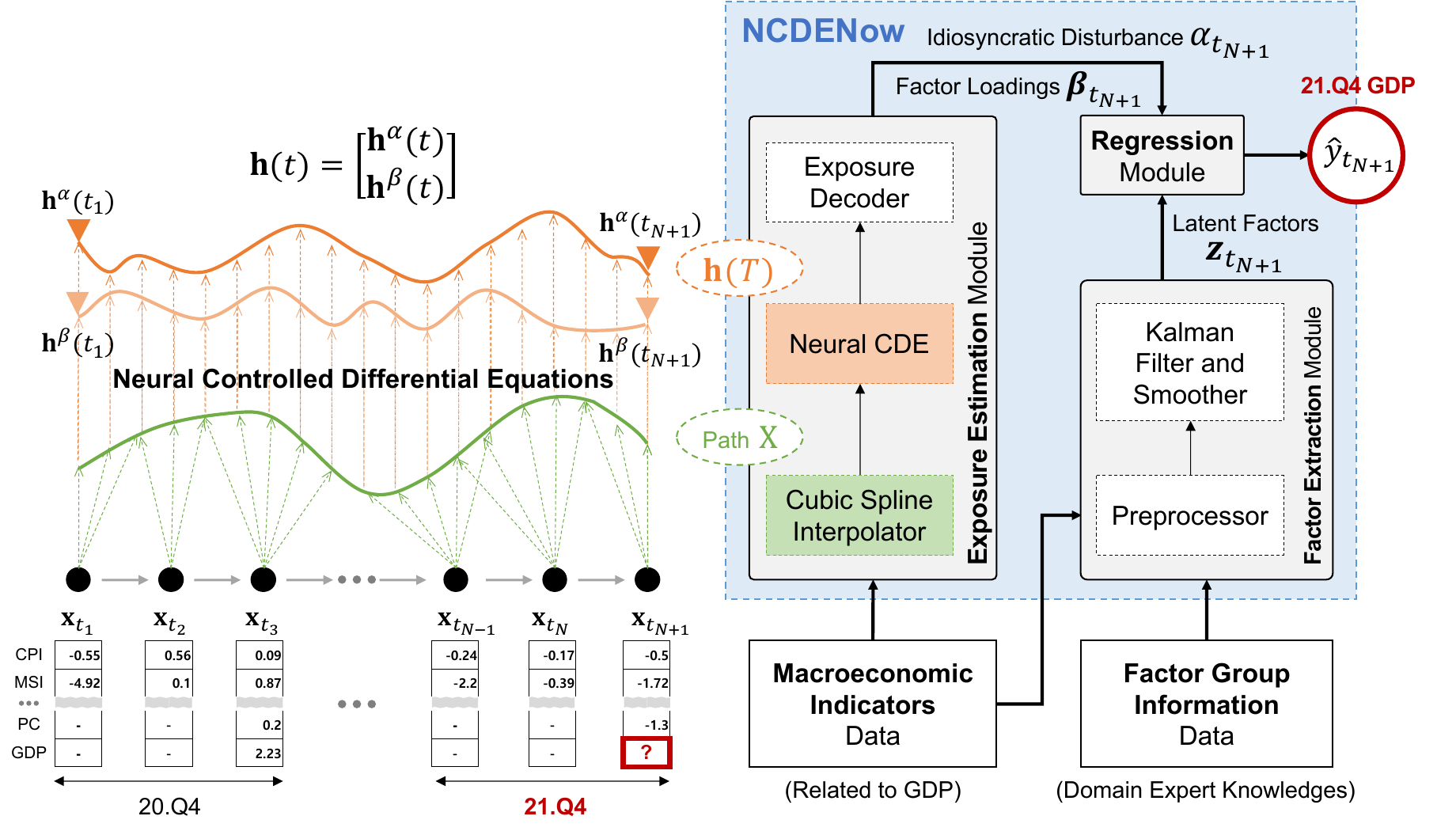}
    \caption{\texttt{NCDENow} workflow illustrating GDP nowcasting for Q4 2021 using macroeconomic indicators from Q4 2020 to Q4 2021.}
    \label{fig:overall_workflow}
\end{figure*}

\subsection{Neural Controlled Differential Equations}
NCDE~\cite{kidger2020neural} is normally regarded as a continuous analogue to RNN and can be written as follows:
\begin{align}\begin{split}
\mathbf{h}(t_b) 
&= \mathbf{h}(t_a) + \int_{t_a}^{t_b} f(\mathbf{h}(t);\mathbf{\theta}_f) dX(t)\\
&= \mathbf{h}(t_a) + \int_{t_a}^{t_b} f(\mathbf{h}(t);\mathbf{\theta}_f) \frac{dX(t)}{dt} dt,\label{eq:ncde}
\end{split}\end{align}
where $f$ is a CDE function, and $\mathbf{h}(t)$ is a hidden vector at time $t$. $X(t)$ is a continuous path created from discrete sequential observations, denoted as $\{(\mathbf{x}_i, t_i)\}_{i=a}^b$.
Especially, NCDE keeps their continuous properties by using the interpolated path $X$ and solving the Riemann-Stieltjes integral to get $\mathbf{h}(t_b)$ from $\mathbf{h}(t_a)$ as shown in Eq.~\eqref{eq:ncde} --- in particular, this problem to derive $\mathbf{h}(t_b)$ from the initial condition $\mathbf{h}(t_a)$ is known as an initial value problem (IVP).
We rely on various ODE solvers to solve the integral problem, such as the explicit Euler method to the 4th order Runge–Kutta (RK4). 
To make the interpolated continuous path $X$, linear interpolation or natural cubic spline interpolation is generally used among several interpolation methods. 
Studies using NCDE are being actively researched in various fields ranging from time-series forecasting to generation~\cite{jhin2021attentive,choi2022graph,jeon2022gt,jhin2022exit,hong2022timekit,jo2023hpcde,lee2023hypernetwork,wi2024continuous,choi2023graph}.

We propose a GDP nowcasting model suitable for uncertain real-world economic conditions. Our model applies the continuous dynamic modeling technique in NCDE to DFM. Moreover, it allows us to analyze the nowcasting results based on factors.

\section{Proposed Method}\label{sec:method}
We design \texttt{NCDENow} based on the DFM and the NCDE. In this section, we first review its overall workflow and then introduce the details.

\subsection{Overall Workflow}
Fig.~\ref{fig:overall_workflow} shows comprehensive designs of our \texttt{NCDENow}. Our method includes exposure estimation, factor extraction, and regression modules. The overall workflow is as follows:
\begin{enumerate}
    \item The exposure estimation module first creates a continuous path $X$ from mixed-frequency macroeconomic time-series $\{\mathbf{F}_i\}^{N}_{i=1}=\{(\mathbf{x}_i,t_i)\}^{N}_{i=1}$, where $\textbf{x}_i\in \mathbb{R}^{D}$ represents $D$ macroeconomic indicators at time $t_i$. We use natural cubic spline interpolation to create the continuous path. This process happens before training our model. Then, the module creates the last hidden vectors for the idiosyncratic disturbance and the factor loading, denoted $\mathbf{h}^{\alpha}(t_N)\in \mathbb{R}^{\dim(\mathbf{h}^{\alpha})}$ and $\mathbf{h}^{\beta}(t_N)\in \mathbb{R}^{\dim(\mathbf{h}^{\beta})}$, respectively.
    Finally, the exposure decoder outputs the scalar value of the idiosyncratic disturbance $\alpha_{t_{N+1}}$ and the vector of factor loading $\mathbf{\bm{\beta}}_{t_{N+1}} \in \mathbb{R}^{K}$, where $K$ is the number of factors.
    The shapes of idiosyncratic disturbance and factor loading differ from the original DFM due to our method's focus on nowcasting GDP, while DFM predicts all economic indicators.
    \item In the pipeline that passes through the factor extraction module, data $\mathbf{F}$ and factor group information provided by economists are input and latent factor $\mathbf{z}_{t_{N+1}}\in \mathbb{R}^{K}$ is output through Kalman filter and smoother.
    \item The final step is the regression module for GDP nowcasting. We use the learned $\alpha_{t_{N+1}}$ and $\mathbf{\bm{\beta}}_{t_{N+1}}$ with the latent factor from the factor extraction modules. Finally, Eq.~\eqref{eq:nowcasting} outputs the predicted GDP growth $\hat{y}_{T_{N+1}}$.
\end{enumerate}

We provide more detailed descriptions for each step in the following subsections.

\subsection{Exposure Estimation Module}\label{subsec:eem}
The exposure estimation module of \texttt{NCDENow} uses cubic spline interpolation to generate continuous paths from economic indicators, preserving the natural dynamics of time series. NCDE learns from these paths to identify idiosyncratic disturbances and factor loadings. A key part of our method is training $\mathbf{\theta}_{f}$ to generate a hidden vector $\mathbf{h}^{\alpha}(t_N)$ of idiosyncratic disturbances and a hidden vector $\mathbf{h}^{\beta}(t_N)$ of factor loadings. To ensure the theoretical accuracy of our proposed method, we share important properties with the path $X$ and at the same time produce better factor loadings and idiosyncratic disturbances. For this purpose, we define the exposure estimation module as an augmented NCDE as follows:
\begin{align} \label{eq:ncde_eq_ours}
        \mathbf{h}(t_N) = \mathbf{h}(t_1) + \int_{t_1}^{t_N} f(\mathbf{h}(t); \mathbf{\theta}_{f})dX(t), \quad
        \mathbf{h}(t_1) = \begin{bmatrix} \mathbf{h}^{\alpha}(t_1) \\ \mathbf{h}^{\beta}(t_1) \end{bmatrix},
\end{align} 
where $\mathbf{h}(t)\in \mathbb{R}^{\dim(\mathbf{h}^{\alpha})+\dim(\mathbf{h}^{\mathbf{\beta}})}$ is the hidden trajectory of the factor loadings and idiosyncratic disturbances, $\mathbf{X}$ is generated through the natural cubic spline from raw observations. 
The initial values $\mathbf{h}^{\alpha}(t_1)$ and $\mathbf{h}^{\beta}(t_1)$ are created from $\mathbf{F}_{t_1}$ as follows:
\begin{align}
    \mathbf{h}^{\alpha}(t_1)&=\texttt{FC}_{D \rightarrow \dim(\mathbf{h}^{\alpha})}(\mathbf{F}_{t_1}), \label{eq:initial_alpha}\\
    \mathbf{h}^{\beta}(t_1)&=\texttt{FC}_{D \rightarrow \dim(\mathbf{h}^{\beta})}(\mathbf{F}_{t_1}),\label{eq:initial_beta}
\end{align}
where $\mathtt{FC}_{input\_size \rightarrow output\_size}$ means a fully-connected layer whose input size is $input\_size$ and output size is also $output\_size$.

The neural network $f(\cdot)$ used in Eq.~\eqref{eq:ncde_eq_ours} is defined as follows:
\begin{align}\begin{split}
f(\mathbf{h}(t);\mathbf{\theta}_f) &= \text{Tanh}(\texttt{FC}_{\dim(\mathbf{h}) \rightarrow dim(\mathbf{h})}(\mathbf{m}_{L})),\\
&\cdots\\
\mathbf{m}_{1} &= \text{ReLU}(\texttt{FC}_{\dim(\mathbf{h}) \rightarrow \dim(\mathbf{h})}(\mathbf{m}_0)),\\
\mathbf{m}_{0} &= \text{ReLU}(\texttt{FC}_{\dim(\mathbf{h}) \rightarrow \dim(\mathbf{h})}(\mathbf{h}(t))), \label{eq:function}
\end{split}\end{align}
which consists of fully-connected layers with ReLU or hyperbolic tangent (Tanh) activation. The size of $\dim(\mathbf{h}$) is the sum of $\dim(\mathbf{h}^{\alpha})$ and $\dim(\mathbf{h}^{\mathbf{\beta}})$. The number of layers $L$ is a hyperparameter.

Let $\mathbf{h}(t_N)$ be the last hidden vector, we have an output layer (i.e., exposure decoder) with a typical construction based on $\mathbf{h}(t_N)$. For the final regression step, the following exposure decoder produces $\alpha_{t_{N+1}}$ and $\mathbf{\bm{\beta}}_{t_{N+1}}$:
\begin{align}
    \alpha_{t_{N+1}} &= \texttt{FC}_{\dim(\mathbf{h}^{\alpha}) \rightarrow 1}(\mathbf{h}^{\alpha}(t_N)), \label{eq:alpha}\\
    \mathbf{\bm{\beta}}_{t_{N+1}} &= \texttt{FC}_{\dim(\mathbf{h}^{\mathbf{\beta}}) \rightarrow K}(\mathbf{h}^{\beta}(t_N)). \label{eq:beta}
\end{align}
This module can incorporate various time series models beyond NCDE, allowing for comparison with RNN-based models in our framework.

\subsection{Factor Extraction Module}
In order to estimate the factors, we use the EM algorithm-based Kalman filtering and smoothing method. The estimated factor loadings ($\mathbf{\Lambda}$) and idiosyncratic disturbances ($\bm{\epsilon}$) in this process are disregarded, as we use the factor loadings ($\mathbf{\bm{\beta}}$) and idiosyncratic disturbance ($\alpha$) re-estimated in the exposure estimation module. The observation and transition equations of DFM are as follows:
\begin{align}\label{eq:tran_obs_eq}
    \begin{split}
        \mathbf{z}_{t_{N+1}} &= \mathbf{A}\mathbf{z}_{t_N} + \bm{\mu}_{t_{N+1}}, \\
        \mathbf{y}_{t_{N+1}} &= \mathbf{\Lambda} \mathbf{z}_{t_{N+1}} + \bm{\epsilon}_{t_{N+1}},
    \end{split}
\end{align} where $\bm{\mu}_{t_{N+1}}$ follows $\mathcal{N}(0, \mathbf{Q})$, $\bm{\epsilon}_{t_{N+1}}$ follows $\mathcal{N}(0, \mathbf{R})$, $\mathbf{Q}$ is the variance of $\bm{\mu}_{t_N}$, and $\mathbf{R}$ is the variance of $\bm{\epsilon}_{t_N}$. The conditional probability distributions of factor $\mathbf{z}$ and observed data $\mathbf{y}$ are as follows:
\begin{align}\label{eq:tran_obs_prob_eq}
    \begin{split}
        P(\mathbf{z}_{t_{N+1}}|\mathbf{z}_{t_N}) &= \mathcal{N}(\mathbf{z}_{t_{N+1}}|\mathbf{A}\mathbf{z}_{t_N},\mathbf{Q}), \\
        P(\mathbf{y}_{t_{N+1}}|\mathbf{z}_{t_{N+1}}) &= \mathcal{N}(\mathbf{y}_{t_{N+1}}|\mathbf{\Lambda} \mathbf{z}_{t_{N+1}},\mathbf{R}).
    \end{split}
\end{align} 

We estimate the parameters of Eqs.~\eqref{eq:tran_obs_eq} and~\eqref{eq:tran_obs_prob_eq} using the expectation-maximization (EM) algorithm~\cite{dempster1977maximum,watson1983alternative,stock2011dynamic}. In the first step, the initial values of parameters $\mathbf{A}, \mathbf{\Lambda}, \mathbf{Q}$, and $\mathbf{R}$ are estimated through principal component analysis (PCA). In the E-step, the factor $\mathbf{z}$ is computed using the Kalman smoothing method, given the observed data and parameters $\mathbf{A}, \mathbf{\Lambda}, \mathbf{Q}$, and $\mathbf{R}$. In the M-step, the parameters $\mathbf{A}, \mathbf{\Lambda}, \mathbf{Q}$, and $\mathbf{R}$ are updated to maximize the expected log-likelihood. 

The factor extraction process is applied independently to groups of correlated indicators, divided based on domain knowledge. This approach follows the method proposed by ~\citet{banbura2010nowcasting}.

\subsection{Regression Module}
The variables $\alpha$ and $\mathbf{\bm{\beta}}$ at time $t_{N+1}$, estimated through exposure estimation module, are regressed with the factor $\mathbf{z}$ extracted through factor extraction module, as shown in Eq.~\eqref{eq:nowcasting}:
\begin{align}\label{eq:nowcasting}
    \hat{y}_{t_{N+1}} = \alpha_{t_{N+1}} + \sum_{k=1}^{K} \beta_{t_{N+1}}^{(k)}z_{t_{N+1}}^{(k)},
\end{align} where $\mathbf{z}$ is factors, $K$ is the number of factors, $\hat{y}_{t_{N+1}}$ is predicted GDP growth at time $t_{N+1}$. 
Eq.~\eqref{eq:nowcasting} is of the same form as the observation equation in Eq.~\eqref{eq:tran_obs_eq}. However, the reason for formulating this equation again is that we re-estimate the variables $\alpha$ and $\mathbf{\bm{\beta}}$ through the exposure estimation module.

\texttt{NCDENow} utilizes dynamic factors extracted from DFM to re-estimate factor loadings and an idiosyncratic disturbance. By doing so, our model employs NCDE to capture the continuous dynamics of macroeconomic indicators. In terms of accuracy, GDP nowcasting in our model outperforms conventional approaches that rely solely on Kalman filtering and smoothing.

\subsection{How to Train}
The overall training algorithm in Alg.~\ref{alg:train}. 
Our model is trained using the following MSE loss function:
\begin{align}\label{eq:loss_mse}
    \mathcal{L} = \frac{\sum_{\tau\in\mathcal{T}} (y_{N+1}^{(\tau)}-\hat{y}_{N+1}^{(\tau)})^{2}}{|\mathcal{T}|},
\end{align}where $\mathcal{T}$ is a training set, $\tau$ is a training sample, $y_{N+1}$ is the ground truth, and $\hat{y}_{N+1}$ is the predicted value.

\begin{algorithm}[t]
    \small
    \SetAlgoLined
    \caption{Training procedure for \texttt{NCDENow}}
    \label{alg:train}
    \KwIn{Training data $\mathcal{T}$, Maximum epochs $max\_epochs$, EM algorithm maximum iterations $max\_iter$, Convergence tolerance $conv\_tol$}
    \KwOut{Trained model parameters}
    \begin{algorithmic}[1]
    \raggedright
    \STATE Initialize all the parameters of \texttt{NCDENow}
    \STATE $epochs \gets 0$

    \WHILE {$epochs < max\_epochs$}
        \STATE Sample a mini-batch $\{\mathbf{x}_i\}_{i=1}^{S} \in \mathcal{T}$\label{a:mini}
    
        \STATE Compute continuous hidden representation $\mathbf{h}(t)$ using NCDE and ODE solver\label{a:cont}

        \STATE $iter \gets 0$
        \STATE $conv\_criterion \gets 0$
    \WHILE {$iter < max\_iter$ \textbf{and} $ conv\_criterion > conv\_tol$}
        \STATE E-step: compute distributions in Eq.~\eqref{eq:tran_obs_prob_eq}
        \STATE M-step: Update parameters to maximize expected log-likelihood
    \ENDWHILE
        \STATE Calculate a $\hat{\mathbf{y}}_{N+1}$ with $\mathbf{\alpha}$, $\mathbf{\bm{\beta}}$, and $\mathbf{z}$
        \STATE Update model parameters using loss function (Eq.~\eqref{eq:loss_mse})
        
        \IF {has not decreased for 5 consecutive epochs} 
        \STATE break
        \ENDIF
        \STATE $epochs \gets epochs + 1$;
    \ENDWHILE
    \STATE \Return Trained model parameters;
\end{algorithmic}
\end{algorithm}

\subsection{Well-posedness of \texttt{NCDENow}}
The concept of well-posedness, which refers to a problem where a solution exists uniquely and alters continuously with variations in input data, has been established for NCDE as shown in \citet[Theorem 1.3]{lyons2004differential} given the Lipschitz continuity condition. Many activation functions, for instance, ReLU, LeakyReLU, SoftPlus, Tanh, Sigmoid, ArcTan, and Softsign, possess a Lipschitz constant value of 1. Other prevalent neural network layers like dropout, batch normalization, and various pooling techniques also have known Lipschitz constants. As such, in certain scenarios of \texttt{NCDENow}, the Lipschitz continuity for $f(\cdot)$ can be achieved, ensuring that the initial value problem represented in Eq.~\eqref{eq:ncde_eq_ours} is well-posed.

\subsection{Properties of \texttt{NCDENow}}
Our key innovation is the integration of DFM with the NCDE framework. \texttt{NCDENow} uses the Kalman filter from DFM flexibly, allowing it to work with different DFMs. Furthermore, our proposed method not only unifies CDEs, but can also be theoretically linked to linear CDEs, including S4~\cite{gu2021efficiently} and Mamba~\cite{gu2023mamba}, which have shown significant advancements in sequence modeling. This enables \texttt{NCDENow} to perform continuous dynamic modeling for lagged or missing data and estimate significant variables in DFM to make accurate nowcasts even under challenging patterns of economic uncertainty.

\section{Experiments} \label{sec:exp}
We conduct experiments with our proposed GDP nowcasting model to address the following research questions:

\begin{itemize}
\item \textbf{RQ1}: Is the performance of our proposed \texttt{NCDENow} superior to the baseline model?
\item \textbf{RQ2}: How is the performance change of \texttt{NCDENow} compared to the baseline according to the missing data rate?
\item \textbf{RQ3}: How well can \texttt{NCDENow} forecast compare to baselines when the volatility of underlying economic data changes?
\item \textbf{RQ4}: How can we interpret the GDP nowcasting results?
\item \textbf{RQ5}: How do different ODE solvers affect the performance of \texttt{NCDENow}?
\item \textbf{RQ6}: How do the different models compare in terms of the number of model parameters?
\end{itemize}

\subsection{Experimental Settings}
\subsubsection{Datasets}

To evaluate \texttt{NCDENow}, we use 2 GDP datasets from South Korea and the United Kingdom.
For South Korea GDP, we collect macroeconomic indicators from Economics Statistics System\footnote{\url{https://ecos.bok.or.kr}}. We include 33 indicators, including 28 months and 5 quarters, collected from Jan. 2002 to Dec. 2021, and following the same setting to \citet{lee2022d}. 

Macroeconomic indicators are categorized into 4 groups: ``Global'', ``Real'', ``Labor'', and ``Soft'', utilized to extract factors: the global group includes all economic indicators; the real group includes indicators related to the real economy; the soft group includes survey data, considering the importance of soft indicators in GDP nowcasting~\cite{kurz2019note}; the labor group consists of indicators related to the labor market. These settings are identical to~\citet{lee2022d} and~\citet{ylee2022a}.

UK GDP dataset is from~\citet{anesti2018uncertain}, which is available for download on the Bank of England website~\footnote{\url{https://www.bankofengland.co.uk/working-paper/2018/uncertain-kingdom-nowcasting-gdp-and-its-revisions}}. Due to the proprietary nature of the series within this dataset, several indicators are not disclosed. 
Consequently, we use 20 publicly available macroeconomic indicators from Oct. 1996 to Dec. 2017, with 12 months and 8 quarters. 
Macroeconomic indicators are categorized into 8 groups and detailed information is included in the dataset we provide.

\subsubsection{Evaluation Metrics}
To evaluate the \texttt{NCDENow} model, we use mean squared error (MSE) and mean absolute percentage error (MAPE). These metrics are as follows:
\begin{align}
    \text{MSE} = \frac{\sum_{s\in\mathcal{S}} (y_{N+1}^{(s)} - \hat{y}_{N+1}^{(s)})^{2}}{|\mathcal{S}|}, 
    \quad
    \text{MAPE} = \frac{1}{|\mathcal{S}|} \sum_{s\in\mathcal{S}} \lvert\frac{y_{N+1}^{(s)} - \hat{y}_{N+1}^{(s)}}{y^{(s)}}\rvert, \nonumber
\end{align}
where $\mathcal{S}$ is a test set, $s$ is a test sample, $y_{N+1}$ is the ground truth, and $\hat{y}_{N+1}$ is the predicted value.

\begin{table}[t]
    \small
    \centering
    \caption{Best hyperparameters of baselines}
    \label{tab:best_baseline}
    \begin{tabular}{ccccccc}\toprule
    \multicolumn{1}{c}{\multirow{2}{*}{Model}} & \multicolumn{3}{c}{Korean GDP} & \multicolumn{3}{c}{UK GDP} \\\cmidrule(lr){2-4} \cmidrule(lr){5-7}
     & lr & \multicolumn{1}{c}{dim($\mathbf{h}$)} & $L$ & lr & \multicolumn{1}{c}{dim($\mathbf{h}$)} &$L$ \\\midrule
    DFM-RNN & \num{1e-2} & 512 & 2 & \num{1e-2} & 256 & 6 \\
    DFM-LSTM & \num{1e-3} & 512 & 4 & \num{1e-3} & 256 & 5 \\
    DFM-GRU & \num{1e-3} & 512 & 5 & \num{1e-2} & 128 & 6 \\
    DFM-NCDE & \num{1e-2} & 128 & 4 & \num{1e-3} & 512 & 5 \\
    NCDE & \num{1e-2} & 128 & 3 & \num{1e-2} & 64 & 4 \\\bottomrule
    \end{tabular}
\end{table}

\begin{table}[t]
    \small
    \setlength{\tabcolsep}{2pt}
    \centering
    \caption{Best hyperparameters of \texttt{NCDENow}}
    \label{tab:best_hyper}
    \begin{tabular}{c cccc cccc}\toprule
    \multicolumn{1}{c}{\multirow{2}{*}{Model}} & \multicolumn{4}{c}{Korean GDP} & \multicolumn{4}{c}{UK GDP} \\\cmidrule(lr){2-5} \cmidrule(lr){6-9}    
    & lr & dim($\mathbf{h}^{\mathbf{\alpha}}$) & dim($\mathbf{h}^{\mathbf{\beta}}$) & $L$ & lr & dim($\mathbf{h}^{\mathbf{\alpha}}$) & dim($\mathbf{h}^{\mathbf{\beta}}$) & $L$\\ \midrule
    \texttt{NCDENow} & \num{1e-2} & 128 & 128 & 1 & \num{1e-3} & 256 & 256  & 1 \\\bottomrule    
    \end{tabular}
\end{table}

\subsubsection{Baselines}
We select RNN-based models such as LSTM~\cite{hochreiter1997long} and GRU~\cite{cho2014properties}, NCDE, and DFM as baselines. Since RNN-based models cannot handle irregular data, they have limitations. Therefore, to transform irregular macroeconomic indicators into regular data, we use DFM as a pre-processor and then conduct training and testing. We consider the following baselines:
\begin{itemize}
\item DFM~\cite{mariano2010coincident,banbura2010nowcasting} conducts GDP nowcasting by utilizing macroeconomic indicators of mixed frequencies.
\item DFM-RNN, DFM-LSTM, DFM-GRU, and DFM-NCDE use DFM for data pre-processing, interpolating quarterly macroeconomic indicators into monthly estimates.
For simplicity, we group DFM-RNN, DFM-LSTM, and DFM-GRU into RNN-based baselines, excluding DFM-NCDE.
\item NCDE~\cite{kidger2020neural} is generally a continuous analogue of RNN, and we use the function of CDE as MLP.
\end{itemize}

\subsubsection{Hyperparameter Settings}
We extract a single factor for each macroeconomic indicator from all baselines using DFM and the exposure estimation module in \texttt{NCDENow}. The order of vector autoregression governing all group dynamics is 1. Each idiosyncratic disturbance for macroeconomic indicators is modeled using a first-order autoregressive process. 
The EM algorithm runs for a maximum of 500 iterations with \num{1e-6} tolerance.

For RNN-based baselines, DFM-NCDE, NCDE, and \texttt{NCDENow}, we train for 1000 epochs using the Adam optimizer, with a batch size of 128 on all datasets. With the validation dataset, an early-stop approach with a patience of 5 iterations is applied. The learning rates are in \{0.01, 0.001, 0.0001\}, the dimensions of $\mathbf{h}^{\mathbf{\alpha}}$ and $\mathbf{h}^{\mathbf{\beta}}$ are in \{64, 128, 256, 512\}, and the number of layers $L$ is in \{1, 2, $\dots$, 6\}. In \texttt{NCDENow} and NCDE, we consider Euler and RK4 as the ODE solver. 

For GDP nowcasting, the look-back window $N=15$. We also list the best hyperparameters for each dataset in Tables~\ref{tab:best_baseline} to ~\ref{tab:best_hyper}. In RNN-based baselines, dim($\mathbf{h}$) represents the dimension of the hidden state and $L$ denotes the number of recurrent layers.

\begin{table}[t]
    \small
    \centering
    \caption{The results of GDP nowcasting. The best results are in \textbf{boldface} and the second-best results are \underline{underlined}.
    `Improv.' indicates the improvement against the best baseline performance.}
    \label{tab:exp_nowcasting}
    \begin{tabular}{ccccc}
    \toprule
    \multirow{2}{*}{Model} & 
    \multicolumn{2}{c}{South Korea GDP} & 
    \multicolumn{2}{c}{UK GDP} \\ \cmidrule(lr){2-3} \cmidrule(lr){4-5}
     & \multicolumn{1}{c}{MSE} & 
     \multicolumn{1}{c}{MAPE} & 
     \multicolumn{1}{c}{MSE} & 
     \multicolumn{1}{c}{MAPE} \\ \midrule
    DFM-RNN & 1.9605 & \underline{0.8676} & 0.0155 & 0.2508 \\
    DFM-LSTM & 1.9715 & 0.9732 & \underline{0.0154} & 0.2515 \\
    DFM-GRU & 1.9763 & 0.9830 & 0.0161 & \underline{0.2381} \\
    DFM-NCDE & 1.9606 & 0.8790 & 0.0163 & 0.2669 \\
    NCDE & 1.9640 & 0.9550 & 0.0155 & 0.2487 \\
    DFM & \underline{1.5044} & 1.1199 & 0.0299 & 0.3396 \\ \midrule
    \texttt{NCDENow} & \textbf{0.5104} & \textbf{0.7049} & \textbf{0.0070} & \textbf{0.1385} \\ \midrule
    Improv. & 66.08\% & 29.11\% & 54.78\% & 41.84\% \\
    \bottomrule
    \end{tabular}
\end{table}

\subsection{Performance Comparison (RQ1)}\label{subsec:rq1}

Table~\ref{tab:exp_nowcasting} shows the experimental results of GDP nowcasting for all models. 
The best results are highlighted in bold. 
\texttt{NCDENow}, a novel framework, outperforms DFM, RNN-based baselines, DFM-NCDE, and NCDE. In particular, for South Korea GDP nowcasting, significant performance improvements of 66.08\% on MSE and 45.08\% on MAPE are achieved compared to DFM.
Similarly, for UK GDP nowcasting, we achieve improved performance with 76.59\% on MSE and 59.22\% on MAPE.

\begin{table}[t]
\small
\centering
\caption{Nowcasting error on missing data}
\label{tab:exp_missing}
\begin{tabular}{ccccc}\toprule
Dataset & Missing Rate & Metrics & DFM & \texttt{NCDENow} \\\midrule
 \multirow{4}{*}{South Korea GDP}  
  & \multirow{2}{*}{10\%} 
  & MSE & 1.7402 & \textbf{0.9438} \\
 && MAPE & 1.2356 & \textbf{0.6039} \\\cmidrule(lr){2-5} 
  & \multirow{2}{*}{20\%} 
  & MSE & 1.3915 & \textbf{1.1887} \\
 && MAPE & 1.0973 & \textbf{0.7979} \\\midrule
 \multirow{4}{*}{UK GDP}
 & \multirow{2}{*}{10\%}
  & MSE & 0.0247 & \textbf{0.0091} \\
 && MAPE    & 0.3000 & \textbf{0.1802} \\\cmidrule(lr){2-5}
 & \multirow{2}{*}{20\%}
 & MSE & 0.0247 & \textbf{0.0103} \\
 && MAPE & 0.3000 & \textbf{0.1815} \\\bottomrule
\end{tabular}
\end{table}

\subsection{Study on Missing Data (RQ2)}\label{subsec:rq2}

\texttt{NCDENow} uses NCDE to estimate factor loadings and idiosyncratic disturbances, handling irregular time series. 
To investigate the impact of increased irregularity on performance, we conduct an experiment by randomly removing 10\% and 20\% of the collected macroeconomic indicators.
The processed dataset is then used to conduct GDP nowcasting using \texttt{NCDENow} and DFM.
In the experimental results Table~\ref{tab:exp_missing}, the performance of \texttt{NCDENow} overperforms that of DFM. Therefore, we conjecture that \texttt{NCDENow} is more robust to nonlinearity caused by missing data compared to DFM.

\subsection{GDP Nowcasting in COVID-19 (RQ3)}\label{subsec:rq3}

We evaluate the effectiveness of \texttt{NCDENow} in capturing sudden economic shifts.
The continuous dynamic modeling capabilities of NCDE allow \texttt{NCDENow} to potentially capture nonlinear patterns and abrupt changes in economic indicators more effectively than DFM.
\begin{wraptable}{r}{0.45\columnwidth}
    \vspace{-1em}
    \small
    \centering
    \caption{Results during the sudden drop in South Korea GDP}
    \label{tab:exp_recession}
    \begin{tabular}{ccc} \toprule
    Model & MSE & MAPE \\ \midrule
    DFM  & 4.6265 & 0.9954 \\
    \texttt{NCDENow} & \textbf{0.9900} & \textbf{0.5387}          \\\bottomrule
    \end{tabular}
    \vspace{-1em}
\end{wraptable}
Table~\ref{tab:exp_recession} shows the result of GDP nowcasting during the initial phase of the COVID-19 pandemic (from Q1 to Q3 2020). As shown in Fig.~\ref{fig:recession}, GDP growth shows a sudden drop during this period. In this context, \texttt{NCDENow} demonstrates superior performance compared to DFM in both MSE and MAPE metrics. Hence, it can be inferred that \texttt{NCDENow} exhibits better capture capabilities of GDP growth during periods of increased economic uncertainty.

\subsection{Factor Analysis (RQ4)}\label{subsec:rq4}

\begin{figure}[h]
    \centering
    \includegraphics[width=0.75\columnwidth]{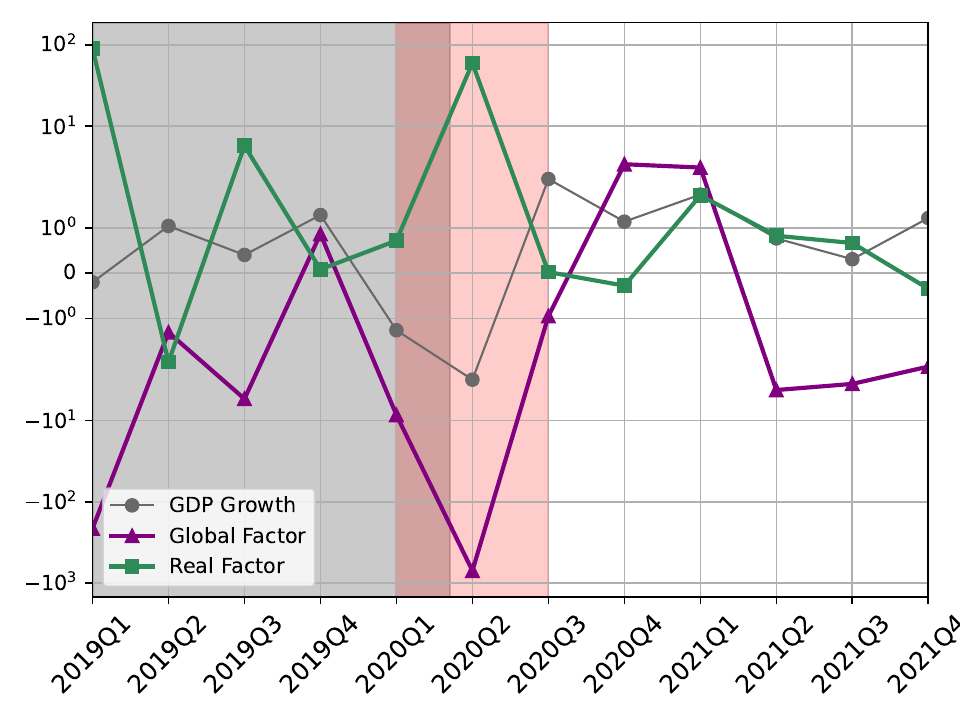}
    \caption{Estimated latent factors from the South Korea GDP dataset. Gray shading indicates the 11th business cycle and COVID-19 economic shock.}
    \label{fig:factors}
\end{figure}

In Fig.~\ref{fig:factors}, we can observe an inverse relationship between the real factor, which reflects real economic activity, and GDP growth. 
Global factors show a strong connection with GDP growth. In the first half of 2020, the global economic downturn and a sharp contraction due to the COVID-19 pandemic closely resemble the changes in global factors. The substantial increase in the real factor during this period is attributed to actions taken in response to COVID-19, including quantitative easing, disaster relief payments, and vaccination efforts.
Furthermore, the rapid increase in global factors in the second half of 2020 can be attributed to effective COVID-19 response measures.

\subsection{Impact of ODE Solvers (RQ5)}\label{subsec:rq5}
We evaluate the influence of ODE solver choice on the performance of \texttt{NCDENow} by comparing the Euler and RK4 methods. RK4 is generally considered more accurate for non-linear differential equations, as it performs four calculations per time step compared to a single Euler computation~\cite{islam2015comparative}. Table~\ref{tab:solver} shows that RK4 consistently outperforms Euler. 

\begin{table}[t!]
\small
\setlength{\tabcolsep}{10pt}
\centering
\caption{Comparison of ODE solvers of \texttt{NCDENow}}
\label{tab:solver}
\begin{tabular}{ccrcr} \toprule
\multirow{2}{*}{ODE Solver} & \multicolumn{2}{c}{South Korea GDP} & \multicolumn{2}{c}{UK GDP} \\\cmidrule(lr){2-3} \cmidrule(lr){4-5}
 & MSE & MAPE & MSE & MAPE \\\midrule
Euler & 1.5470 & 1.3629 & 0.0988 & 0.5553 \\
RK4 & \textbf{0.5104} & \textbf{0.7049} & \textbf{0.0070} & \textbf{0.1385} \\\bottomrule
\end{tabular}
\end{table}

\subsection{Model Efficiency Analyses (RQ6)}\label{subsec:rq6}
\begin{figure}[t]
    \centering
    \subfigure[South Korea GDP]{\includegraphics[width=0.49\columnwidth]{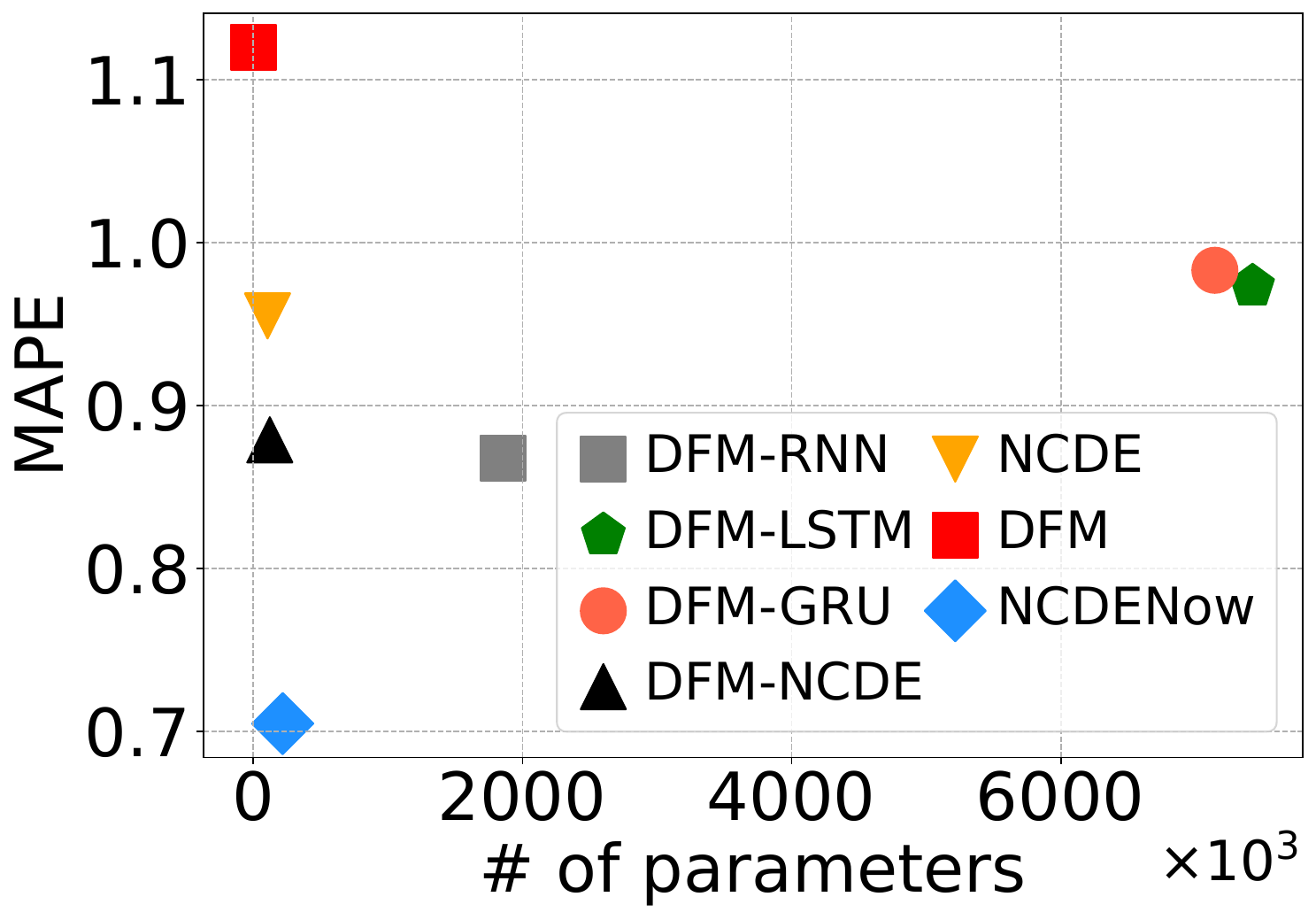}}
    \subfigure[UK GDP]{\includegraphics[width=0.49\columnwidth]{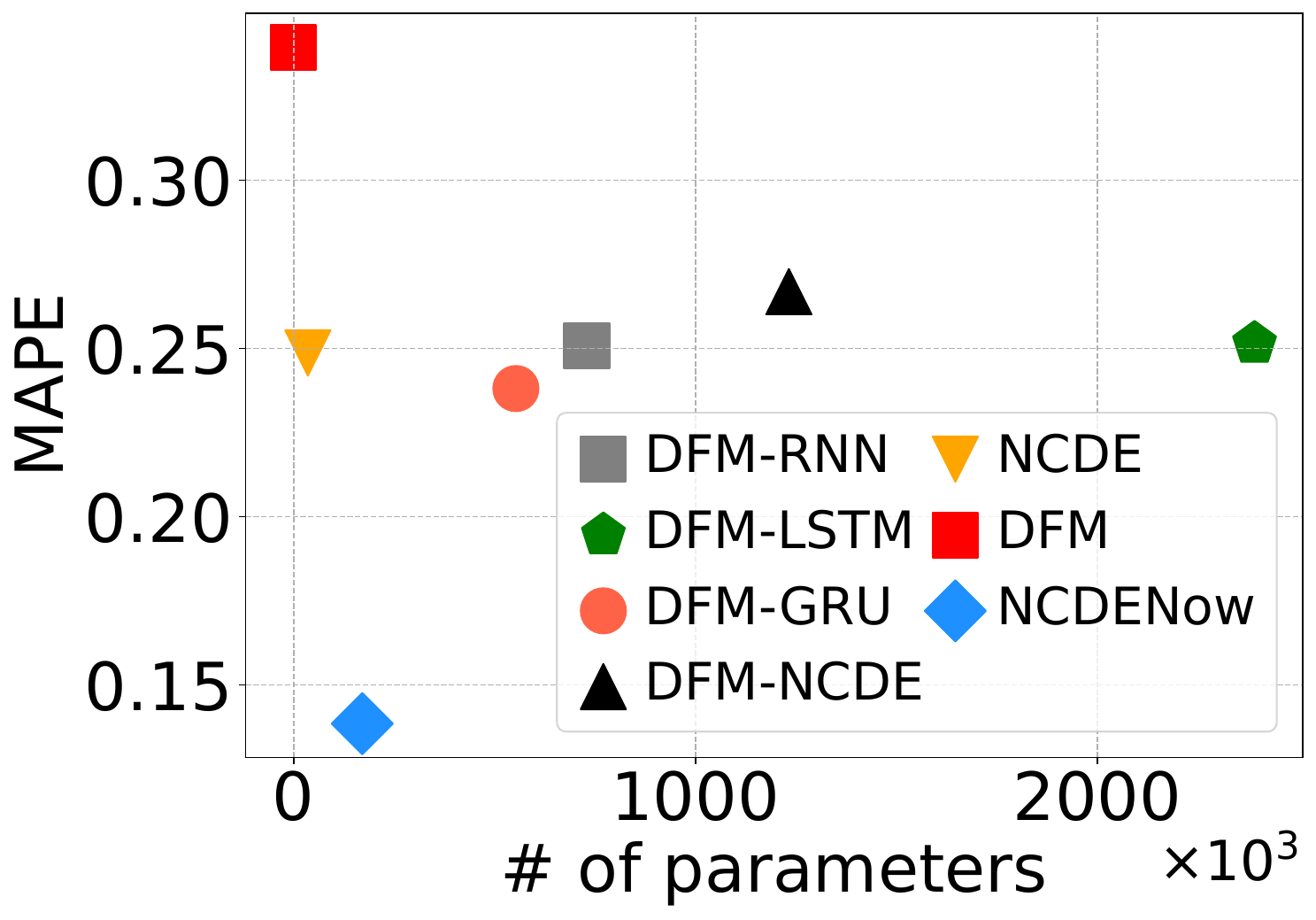}}
    \caption{MAPE versus number of parameters}
    \label{fig:param}
\end{figure}

Fig.~\ref{fig:param} shows the number of parameters and the MAPE. Models in the bottom left corner of this figure are preferred. Our \texttt{NCDENow} models are located around the bottom left. On both datasets, RNN-based baselines and DFM-NCDE show low efficiencies. NCDE is located in the upper left corner and has few parameters but low performance.
For South Korea GDP nowcasting, \texttt{NCDENow} uses about 4\% of the average parameters of DFM-RNN but outperforms it by approximately 3.86 times in MSE. With about twice as many parameters as NCDE, \texttt{NCDENow} shows a performance improvement of approximately 3.85 times in MSE.

\section{Conclusion and Limitation}

We introduced a novel model, \texttt{NCDENow}, that combines the strengths of DFM and NCDE for GDP nowcasting.
Designed to handle sudden drops and irregular data,  \texttt{NCDENow} has been tested under various scenarios, such as different rates of missing data and data volatility.
The findings reveal that \texttt{NCDENow} outperforms 6 baselines in accuracy. Furthermore, the analysis of factors extracted by our model provides valuable insights for economic policy-making.

Our experiments focusing only on 2 datasets might limit the applicability and generalizability of \texttt{NCDENow} in global economic contexts.
We will broaden the range of datasets for future work and incorporate continuous modeling of the parameter estimation process within Kalman filtering and smoothing.

\section*{Acknowledgements}
This work was partly supported by Institute for Information \& Communications Technology Planning \& Evaluation (IITP) grants funded by the Korea government (MSIT) (No. RS-2020-II201361, Artificial Intelligence Graduate School Program (Yonsei University), 10\%), (No.2022-0-00857, Development of Al/data-based financial/economic digital twin platform, 45\%) and (No.RS-2022-II220113, Developing a Sustainable Collaborative Multi-modal Lifelong Learning Framework, 45\%)

\clearpage
\onecolumn
\begin{multicols}{2}
    \bibliographystyle{ACM-Reference-Format}
    \bibliography{NCDENow}


\begin{thebibliography}{45}


\ifx \showCODEN    \undefined \def \showCODEN     #1{\unskip}     \fi
\ifx \showDOI      \undefined \def \showDOI       #1{#1}\fi
\ifx \showISBNx    \undefined \def \showISBNx     #1{\unskip}     \fi
\ifx \showISBNxiii \undefined \def \showISBNxiii  #1{\unskip}     \fi
\ifx \showISSN     \undefined \def \showISSN      #1{\unskip}     \fi
\ifx \showLCCN     \undefined \def \showLCCN      #1{\unskip}     \fi
\ifx \shownote     \undefined \def \shownote      #1{#1}          \fi
\ifx \showarticletitle \undefined \def \showarticletitle #1{#1}   \fi
\ifx \showURL      \undefined \def \showURL       {\relax}        \fi
\providecommand\bibfield[2]{#2}
\providecommand\bibinfo[2]{#2}
\providecommand\natexlab[1]{#1}
\providecommand\showeprint[2][]{arXiv:#2}

\bibitem[Almuzara et~al\mbox{.}(2023)]%
        {almuzara2023new}
\bibfield{author}{\bibinfo{person}{Martin Almuzara}, \bibinfo{person}{Katie Baker}, \bibinfo{person}{Hannah O’Keeffe}, {and} \bibinfo{person}{Argia Sbordone}.} \bibinfo{year}{2023}\natexlab{}.
\newblock \showarticletitle{The New York Fed Staff Nowcast 2.0}.
\newblock  (\bibinfo{year}{2023}).
\newblock


\bibitem[Anesti et~al\mbox{.}(2018)]%
        {anesti2018uncertain}
\bibfield{author}{\bibinfo{person}{Nikoleta Anesti}, \bibinfo{person}{Ana~Beatriz Galv{\~a}o}, {and} \bibinfo{person}{Silvia Miranda-Agrippino}.} \bibinfo{year}{2018}\natexlab{}.
\newblock \showarticletitle{Uncertain Kingdom: nowcasting GDP and its revisions}.
\newblock  (\bibinfo{year}{2018}).
\newblock


\bibitem[Ashwin et~al\mbox{.}(2021)]%
        {ashwin2021nowcasting}
\bibfield{author}{\bibinfo{person}{Julian Ashwin}, \bibinfo{person}{Eleni Kalamara}, {and} \bibinfo{person}{Lorena Saiz}.} \bibinfo{year}{2021}\natexlab{}.
\newblock \showarticletitle{Nowcasting euro area GDP with news sentiment: a tale of two crises}.
\newblock  (\bibinfo{year}{2021}).
\newblock


\bibitem[Assun{\c{c}}ao and Fernandes(2022)]%
        {assunccao2022nowcasting}
\bibfield{author}{\bibinfo{person}{Joao~B Assun{\c{c}}ao} {and} \bibinfo{person}{Pedro~Afonso Fernandes}.} \bibinfo{year}{2022}\natexlab{}.
\newblock \showarticletitle{Nowcasting the Portuguese GDP with Monthly Data}.
\newblock \bibinfo{journal}{\emph{arXiv preprint arXiv:2206.06823}} (\bibinfo{year}{2022}).
\newblock


\bibitem[Baker et~al\mbox{.}(2023)]%
        {baker2023reintroducing}
\bibfield{author}{\bibinfo{person}{Katie Baker}, \bibinfo{person}{Mart{\'\i}n Almuzara}, \bibinfo{person}{Hannah O’Keeffe}, {and} \bibinfo{person}{Argia~M Sbordone}.} \bibinfo{year}{2023}\natexlab{}.
\newblock \bibinfo{booktitle}{\emph{Reintroducing the New York Fed Staff Nowcast}}.
\newblock \bibinfo{type}{{T}echnical {R}eport}. \bibinfo{institution}{Federal Reserve Bank of New York}.
\newblock


\bibitem[Ba{\'n}bura et~al\mbox{.}(2013)]%
        {banbura2013now}
\bibfield{author}{\bibinfo{person}{Marta Ba{\'n}bura}, \bibinfo{person}{Domenico Giannone}, \bibinfo{person}{Michele Modugno}, {and} \bibinfo{person}{Lucrezia Reichlin}.} \bibinfo{year}{2013}\natexlab{}.
\newblock \showarticletitle{Now-casting and the real-time data flow}.
\newblock In \bibinfo{booktitle}{\emph{Handbook of economic forecasting}}. Vol.~\bibinfo{volume}{2}. \bibinfo{publisher}{Elsevier}, \bibinfo{pages}{195--237}.
\newblock


\bibitem[Banbura et~al\mbox{.}(2010)]%
        {banbura2010nowcasting}
\bibfield{author}{\bibinfo{person}{Marta Banbura}, \bibinfo{person}{Domenico Giannone}, {and} \bibinfo{person}{Lucrezia Reichlin}.} \bibinfo{year}{2010}\natexlab{}.
\newblock \showarticletitle{Nowcasting}.
\newblock  (\bibinfo{year}{2010}).
\newblock


\bibitem[Blazquez-Soriano and Ramos-Sandoval({[n.\,d.]})]%
        {blazqueznowcasting}
\bibfield{author}{\bibinfo{person}{Amparo Blazquez-Soriano} {and} \bibinfo{person}{Rosmery Ramos-Sandoval}.} \bibinfo{year}{[n.\,d.]}\natexlab{}.
\newblock \showarticletitle{Nowcasting and forecasting with Big Data}.
\newblock In \bibinfo{booktitle}{\emph{Elgar Encyclopedia of Technology and Politics}}. \bibinfo{publisher}{Edward Elgar Publishing}, \bibinfo{pages}{34--37}.
\newblock


\bibitem[Bontempi et~al\mbox{.}(2017)]%
        {bontempi2017dynamic}
\bibfield{author}{\bibinfo{person}{Gianluca Bontempi}, \bibinfo{person}{Yann-A{\"e}l Le~Borgne}, {and} \bibinfo{person}{Jacopo De~Stefani}.} \bibinfo{year}{2017}\natexlab{}.
\newblock \showarticletitle{A dynamic factor machine learning method for multi-variate and multi-step-ahead forecasting}. In \bibinfo{booktitle}{\emph{2017 IEEE International Conference on Data Science and Advanced Analytics (DSAA)}}. IEEE, \bibinfo{pages}{222--231}.
\newblock


\bibitem[Cho et~al\mbox{.}(2014)]%
        {cho2014properties}
\bibfield{author}{\bibinfo{person}{Kyunghyun Cho}, \bibinfo{person}{Bart Van~Merri{\"e}nboer}, \bibinfo{person}{Dzmitry Bahdanau}, {and} \bibinfo{person}{Yoshua Bengio}.} \bibinfo{year}{2014}\natexlab{}.
\newblock \showarticletitle{On the properties of neural machine translation: Encoder-decoder approaches}.
\newblock \bibinfo{journal}{\emph{arXiv preprint arXiv:1409.1259}} (\bibinfo{year}{2014}).
\newblock


\bibitem[Choi et~al\mbox{.}(2022)]%
        {choi2022graph}
\bibfield{author}{\bibinfo{person}{Jeongwhan Choi}, \bibinfo{person}{Hwangyong Choi}, \bibinfo{person}{Jeehyun Hwang}, {and} \bibinfo{person}{Noseong Park}.} \bibinfo{year}{2022}\natexlab{}.
\newblock \showarticletitle{Graph neural controlled differential equations for traffic forecasting}. In \bibinfo{booktitle}{\emph{Proceedings of the AAAI Conference on Artificial Intelligence}}, Vol.~\bibinfo{volume}{36}. \bibinfo{pages}{6367--6374}.
\newblock


\bibitem[Choi and Park(2023)]%
        {choi2023graph}
\bibfield{author}{\bibinfo{person}{Jeongwhan Choi} {and} \bibinfo{person}{Noseong Park}.} \bibinfo{year}{2023}\natexlab{}.
\newblock \showarticletitle{Graph neural rough differential equations for traffic forecasting}.
\newblock \bibinfo{journal}{\emph{ACM Transactions on Intelligent Systems and Technology}} \bibinfo{volume}{14}, \bibinfo{number}{4} (\bibinfo{year}{2023}), \bibinfo{pages}{1--27}.
\newblock


\bibitem[Cirone et~al\mbox{.}(2024)]%
        {cirone2024theoretical}
\bibfield{author}{\bibinfo{person}{Nicola~Muca Cirone}, \bibinfo{person}{Antonio Orvieto}, \bibinfo{person}{Benjamin Walker}, \bibinfo{person}{Cristopher Salvi}, {and} \bibinfo{person}{Terry Lyons}.} \bibinfo{year}{2024}\natexlab{}.
\newblock \showarticletitle{Theoretical Foundations of Deep Selective State-Space Models}.
\newblock \bibinfo{journal}{\emph{arXiv preprint arXiv:2402.19047}} (\bibinfo{year}{2024}).
\newblock


\bibitem[Cohen et~al\mbox{.}(2023)]%
        {cohen2023nowcasting}
\bibfield{author}{\bibinfo{person}{Samuel~N Cohen}, \bibinfo{person}{Silvia Lui}, \bibinfo{person}{Will Malpass}, \bibinfo{person}{Giulia Mantoan}, \bibinfo{person}{Lars Nesheim}, \bibinfo{person}{Aureo de Paula}, \bibinfo{person}{Andrew Reeves}, \bibinfo{person}{Craig Scott}, \bibinfo{person}{Emma Small}, {and} \bibinfo{person}{Lingyi Yang}.} \bibinfo{year}{2023}\natexlab{}.
\newblock \showarticletitle{Nowcasting with signature methods}.
\newblock \bibinfo{journal}{\emph{arXiv preprint arXiv:2305.10256}} (\bibinfo{year}{2023}).
\newblock


\bibitem[Dempster et~al\mbox{.}(1977)]%
        {dempster1977maximum}
\bibfield{author}{\bibinfo{person}{Arthur~P Dempster}, \bibinfo{person}{Nan~M Laird}, {and} \bibinfo{person}{Donald~B Rubin}.} \bibinfo{year}{1977}\natexlab{}.
\newblock \showarticletitle{Maximum likelihood from incomplete data via the EM algorithm}.
\newblock \bibinfo{journal}{\emph{Journal of the royal statistical society: series B (methodological)}} \bibinfo{volume}{39}, \bibinfo{number}{1} (\bibinfo{year}{1977}), \bibinfo{pages}{1--22}.
\newblock


\bibitem[Duan et~al\mbox{.}(2022)]%
        {duan2022factorvae}
\bibfield{author}{\bibinfo{person}{Yitong Duan}, \bibinfo{person}{Lei Wang}, \bibinfo{person}{Qizhong Zhang}, {and} \bibinfo{person}{Jian Li}.} \bibinfo{year}{2022}\natexlab{}.
\newblock \showarticletitle{FactorVAE: A Probabilistic Dynamic Factor Model Based on Variational Autoencoder for Predicting Cross-Sectional Stock Returns}. In \bibinfo{booktitle}{\emph{Proceedings of the AAAI Conference on Artificial Intelligence}}, Vol.~\bibinfo{volume}{36}. \bibinfo{pages}{4468--4476}.
\newblock


\bibitem[Geweke(1977)]%
        {geweke1977dynamic}
\bibfield{author}{\bibinfo{person}{John Geweke}.} \bibinfo{year}{1977}\natexlab{}.
\newblock \showarticletitle{The dynamic factor analysis of economic time series}.
\newblock \bibinfo{journal}{\emph{Latent variables in socio-economic models}} (\bibinfo{year}{1977}).
\newblock


\bibitem[Giannone et~al\mbox{.}(2008)]%
        {giannone2008nowcasting}
\bibfield{author}{\bibinfo{person}{Domenico Giannone}, \bibinfo{person}{Lucrezia Reichlin}, {and} \bibinfo{person}{David Small}.} \bibinfo{year}{2008}\natexlab{}.
\newblock \showarticletitle{Nowcasting: The real-time informational content of macroeconomic data}.
\newblock \bibinfo{journal}{\emph{Journal of monetary economics}} \bibinfo{volume}{55}, \bibinfo{number}{4} (\bibinfo{year}{2008}), \bibinfo{pages}{665--676}.
\newblock


\bibitem[Gu and Dao(2023)]%
        {gu2023mamba}
\bibfield{author}{\bibinfo{person}{Albert Gu} {and} \bibinfo{person}{Tri Dao}.} \bibinfo{year}{2023}\natexlab{}.
\newblock \showarticletitle{Mamba: Linear-time sequence modeling with selective state spaces}.
\newblock \bibinfo{journal}{\emph{arXiv preprint arXiv:2312.00752}} (\bibinfo{year}{2023}).
\newblock


\bibitem[Gu et~al\mbox{.}(2021)]%
        {gu2021efficiently}
\bibfield{author}{\bibinfo{person}{Albert Gu}, \bibinfo{person}{Karan Goel}, {and} \bibinfo{person}{Christopher R{\'e}}.} \bibinfo{year}{2021}\natexlab{}.
\newblock \showarticletitle{Efficiently modeling long sequences with structured state spaces}.
\newblock \bibinfo{journal}{\emph{arXiv preprint arXiv:2111.00396}} (\bibinfo{year}{2021}).
\newblock


\bibitem[Hayashi and Tachi(2023)]%
        {hayashi2023nowcasting}
\bibfield{author}{\bibinfo{person}{Fumio Hayashi} {and} \bibinfo{person}{Yuta Tachi}.} \bibinfo{year}{2023}\natexlab{}.
\newblock \showarticletitle{Nowcasting japan’s gdp}.
\newblock \bibinfo{journal}{\emph{Empirical Economics}} \bibinfo{volume}{64}, \bibinfo{number}{4} (\bibinfo{year}{2023}), \bibinfo{pages}{1699--1735}.
\newblock


\bibitem[Higgins(2014)]%
        {higgins2014gdpnow}
\bibfield{author}{\bibinfo{person}{Patrick~C Higgins}.} \bibinfo{year}{2014}\natexlab{}.
\newblock \showarticletitle{GDPNow: A Model for GDP'Nowcasting'}.
\newblock  (\bibinfo{year}{2014}).
\newblock


\bibitem[Hochreiter and Schmidhuber(1997)]%
        {hochreiter1997long}
\bibfield{author}{\bibinfo{person}{Sepp Hochreiter} {and} \bibinfo{person}{J{\"u}rgen Schmidhuber}.} \bibinfo{year}{1997}\natexlab{}.
\newblock \showarticletitle{Long short-term memory}.
\newblock \bibinfo{journal}{\emph{Neural computation}} \bibinfo{volume}{9}, \bibinfo{number}{8} (\bibinfo{year}{1997}), \bibinfo{pages}{1735--1780}.
\newblock


\bibitem[Hong et~al\mbox{.}(2022)]%
        {hong2022timekit}
\bibfield{author}{\bibinfo{person}{Seoyoung Hong}, \bibinfo{person}{Minju Jo}, \bibinfo{person}{Seungji Kook}, \bibinfo{person}{Jaeeun Jung}, \bibinfo{person}{Hyowon Wi}, \bibinfo{person}{Noseong Park}, {and} \bibinfo{person}{Sung-Bae Cho}.} \bibinfo{year}{2022}\natexlab{}.
\newblock \showarticletitle{{TimeKit}: A Time-series Forecasting-based Upgrade Kit for Collaborative Filtering}. In \bibinfo{booktitle}{\emph{2022 IEEE International Conference on Big Data (Big Data)}}. IEEE, \bibinfo{pages}{565--574}.
\newblock


\bibitem[Islam(2015)]%
        {islam2015comparative}
\bibfield{author}{\bibinfo{person}{Md~Amirul Islam}.} \bibinfo{year}{2015}\natexlab{}.
\newblock \showarticletitle{A comparative study on numerical solutions of initial value problems (IVP) for ordinary differential equations (ODE) with Euler and Runge Kutta Methods}.
\newblock \bibinfo{journal}{\emph{American Journal of computational mathematics}} \bibinfo{volume}{5}, \bibinfo{number}{03} (\bibinfo{year}{2015}), \bibinfo{pages}{393--404}.
\newblock


\bibitem[Jardet and Meunier(2022)]%
        {jardet2022nowcasting}
\bibfield{author}{\bibinfo{person}{Caroline Jardet} {and} \bibinfo{person}{Baptiste Meunier}.} \bibinfo{year}{2022}\natexlab{}.
\newblock \showarticletitle{Nowcasting world GDP growth with high-frequency data}.
\newblock \bibinfo{journal}{\emph{Journal of Forecasting}} \bibinfo{volume}{41}, \bibinfo{number}{6} (\bibinfo{year}{2022}), \bibinfo{pages}{1181--1200}.
\newblock


\bibitem[Jeon et~al\mbox{.}(2022)]%
        {jeon2022gt}
\bibfield{author}{\bibinfo{person}{Jinsung Jeon}, \bibinfo{person}{Jeonghak Kim}, \bibinfo{person}{Haryong Song}, \bibinfo{person}{Seunghyeon Cho}, {and} \bibinfo{person}{Noseong Park}.} \bibinfo{year}{2022}\natexlab{}.
\newblock \showarticletitle{GT-GAN: General Purpose Time Series Synthesis with Generative Adversarial Networks}.
\newblock \bibinfo{journal}{\emph{Advances in Neural Information Processing Systems}}  \bibinfo{volume}{35} (\bibinfo{year}{2022}), \bibinfo{pages}{36999--37010}.
\newblock


\bibitem[Jhin et~al\mbox{.}(2022)]%
        {jhin2022exit}
\bibfield{author}{\bibinfo{person}{Sheo~Yon Jhin}, \bibinfo{person}{Jaehoon Lee}, \bibinfo{person}{Minju Jo}, \bibinfo{person}{Seungji Kook}, \bibinfo{person}{Jinsung Jeon}, \bibinfo{person}{Jihyeon Hyeong}, \bibinfo{person}{Jayoung Kim}, {and} \bibinfo{person}{Noseong Park}.} \bibinfo{year}{2022}\natexlab{}.
\newblock \showarticletitle{Exit: Extrapolation and interpolation-based neural controlled differential equations for time-series classification and forecasting}. In \bibinfo{booktitle}{\emph{Proceedings of the ACM Web Conference 2022}}. \bibinfo{pages}{3102--3112}.
\newblock


\bibitem[Jhin et~al\mbox{.}(2021)]%
        {jhin2021attentive}
\bibfield{author}{\bibinfo{person}{Sheo~Yon Jhin}, \bibinfo{person}{Heejoo Shin}, \bibinfo{person}{Seoyoung Hong}, \bibinfo{person}{Minju Jo}, \bibinfo{person}{Solhee Park}, \bibinfo{person}{Noseong Park}, \bibinfo{person}{Seungbeom Lee}, \bibinfo{person}{Hwiyoung Maeng}, {and} \bibinfo{person}{Seungmin Jeon}.} \bibinfo{year}{2021}\natexlab{}.
\newblock \showarticletitle{Attentive neural controlled differential equations for time-series classification and forecasting}. In \bibinfo{booktitle}{\emph{2021 IEEE International Conference on Data Mining (ICDM)}}. IEEE, \bibinfo{pages}{250--259}.
\newblock


\bibitem[Jo et~al\mbox{.}(2023)]%
        {jo2023hpcde}
\bibfield{author}{\bibinfo{person}{Minju Jo}, \bibinfo{person}{Seungji Kook}, {and} \bibinfo{person}{Noseong Park}.} \bibinfo{year}{2023}\natexlab{}.
\newblock \showarticletitle{Hawkes Process Based on Controlled Differential Equations}. In \bibinfo{booktitle}{\emph{Proceedings of the Thirty-Second International Joint Conference on Artificial Intelligence (IJCAI)}}. \bibinfo{publisher}{International Joint Conferences on Artificial Intelligence Organization}, \bibinfo{pages}{2151--2159}.
\newblock


\bibitem[Kidger et~al\mbox{.}(2020)]%
        {kidger2020neural}
\bibfield{author}{\bibinfo{person}{Patrick Kidger}, \bibinfo{person}{James Morrill}, \bibinfo{person}{James Foster}, {and} \bibinfo{person}{Terry Lyons}.} \bibinfo{year}{2020}\natexlab{}.
\newblock \showarticletitle{Neural controlled differential equations for irregular time series}.
\newblock \bibinfo{journal}{\emph{Advances in Neural Information Processing Systems}}  \bibinfo{volume}{33} (\bibinfo{year}{2020}), \bibinfo{pages}{6696--6707}.
\newblock


\bibitem[Kurz-Kim(2019)]%
        {kurz2019note}
\bibfield{author}{\bibinfo{person}{Jeong-Ryeol Kurz-Kim}.} \bibinfo{year}{2019}\natexlab{}.
\newblock \showarticletitle{A note on the predictive power of survey data in nowcasting euro area GDP}.
\newblock \bibinfo{journal}{\emph{Journal of Forecasting}} \bibinfo{volume}{38}, \bibinfo{number}{6} (\bibinfo{year}{2019}), \bibinfo{pages}{489--503}.
\newblock


\bibitem[Lee et~al\mbox{.}({[n.\,d.]})]%
        {lee2023hypernetwork}
\bibfield{author}{\bibinfo{person}{Jaehoon Lee}, \bibinfo{person}{Chan Kim}, \bibinfo{person}{Gyumin Lee}, \bibinfo{person}{Haksoo Lim}, \bibinfo{person}{Jeongwhan Choi}, \bibinfo{person}{Kookjin Lee}, \bibinfo{person}{Dongeun Lee}, \bibinfo{person}{Sanghyun Hong}, {and} \bibinfo{person}{Noseong Park}.} \bibinfo{year}{[n.\,d.]}\natexlab{}.
\newblock \showarticletitle{HyperNetwork Approximating Future Parameters for Time Series Forecasting under Temporal Drifts}. In \bibinfo{booktitle}{\emph{NeurIPS 2023 Workshop on Distribution Shifts: New Frontiers with Foundation Models}}.
\newblock


\bibitem[Luo and Yu(2021)]%
        {luo2021determining}
\bibfield{author}{\bibinfo{person}{Jiayi Luo} {and} \bibinfo{person}{Cindy~Long Yu}.} \bibinfo{year}{2021}\natexlab{}.
\newblock \showarticletitle{Determining number of factors in dynamic factor models contributing to GDP nowcasting}.
\newblock \bibinfo{journal}{\emph{Mathematics}} \bibinfo{volume}{9}, \bibinfo{number}{22} (\bibinfo{year}{2021}), \bibinfo{pages}{2865}.
\newblock


\bibitem[Lyons et~al\mbox{.}(2004)]%
        {lyons2004differential}
\bibfield{author}{\bibinfo{person}{Terry Lyons}, \bibinfo{person}{M. Caruana}, {and} \bibinfo{person}{T. Lévy}.} \bibinfo{year}{2004}\natexlab{}.
\newblock \bibinfo{booktitle}{\emph{Differential Equations Driven by Rough Paths}}.
\newblock \bibinfo{publisher}{Springer}.
\newblock
\newblock
\shownote{École D'Eté de Probabilités de Saint-Flour XXXIV - 2004}.


\bibitem[Mariano and Murasawa(2010)]%
        {mariano2010coincident}
\bibfield{author}{\bibinfo{person}{Roberto~S Mariano} {and} \bibinfo{person}{Yasutomo Murasawa}.} \bibinfo{year}{2010}\natexlab{}.
\newblock \showarticletitle{A coincident index, common factors, and monthly real GDP}.
\newblock \bibinfo{journal}{\emph{Oxford Bulletin of economics and statistics}} \bibinfo{volume}{72}, \bibinfo{number}{1} (\bibinfo{year}{2010}), \bibinfo{pages}{27--46}.
\newblock


\bibitem[N{\'e}meth and Hadh{\'a}zi(2023)]%
        {nemeth2023gdp}
\bibfield{author}{\bibinfo{person}{Krist{\'o}f N{\'e}meth} {and} \bibinfo{person}{D{\'a}niel Hadh{\'a}zi}.} \bibinfo{year}{2023}\natexlab{}.
\newblock \showarticletitle{GDP nowcasting with artificial neural networks: How much does long-term memory matter?}
\newblock \bibinfo{journal}{\emph{arXiv preprint arXiv:2304.05805}} (\bibinfo{year}{2023}).
\newblock


\bibitem[S.~Lee(2017)]%
        {lee2017s}
\bibfield{author}{\bibinfo{person}{J.~Han S.~Lee, E.~Lee}.} \bibinfo{year}{2017}\natexlab{}.
\newblock \showarticletitle{Short-term forecasting system using machine learning and mixed cycle model}.
\newblock \bibinfo{journal}{\emph{Bank of Korea Articles in Monthly Bulletin}} (\bibinfo{year}{2017}).
\newblock


\bibitem[Stock and Watson(2011)]%
        {stock2011dynamic}
\bibfield{author}{\bibinfo{person}{James~H Stock} {and} \bibinfo{person}{Mark~W Watson}.} \bibinfo{year}{2011}\natexlab{}.
\newblock \showarticletitle{Dynamic factor models}.
\newblock  (\bibinfo{year}{2011}).
\newblock


\bibitem[Watson and Engle(1983)]%
        {watson1983alternative}
\bibfield{author}{\bibinfo{person}{Mark~W Watson} {and} \bibinfo{person}{Robert~F Engle}.} \bibinfo{year}{1983}\natexlab{}.
\newblock \showarticletitle{Alternative algorithms for the estimation of dynamic factor, mimic and varying coefficient regression models}.
\newblock \bibinfo{journal}{\emph{Journal of Econometrics}} \bibinfo{volume}{23}, \bibinfo{number}{3} (\bibinfo{year}{1983}), \bibinfo{pages}{385--400}.
\newblock


\bibitem[Wi et~al\mbox{.}(2024)]%
        {wi2024continuous}
\bibfield{author}{\bibinfo{person}{Hyowon Wi}, \bibinfo{person}{Yehjin Shin}, {and} \bibinfo{person}{Noseong Park}.} \bibinfo{year}{2024}\natexlab{}.
\newblock \showarticletitle{Continuous-time Autoencoders for Regular and Irregular Time Series Imputation}. In \bibinfo{booktitle}{\emph{Proceedings of the 17th ACM International Conference on Web Search and Data Mining}}. \bibinfo{pages}{826--835}.
\newblock


\bibitem[Woloszko(2020)]%
        {woloszko2020adaptive}
\bibfield{author}{\bibinfo{person}{Nicolas Woloszko}.} \bibinfo{year}{2020}\natexlab{}.
\newblock \showarticletitle{Adaptive Trees: a new approach to economic forecasting}.
\newblock  (\bibinfo{year}{2020}).
\newblock


\bibitem[Wu et~al\mbox{.}(2021)]%
        {wu2021data}
\bibfield{author}{\bibinfo{person}{Xin Wu}, \bibinfo{person}{Zhenyuan Zhang}, \bibinfo{person}{Haotian Chang}, {and} \bibinfo{person}{Qi Huang}.} \bibinfo{year}{2021}\natexlab{}.
\newblock \showarticletitle{A data-driven gross domestic product forecasting model based on multi-indicator assessment}.
\newblock \bibinfo{journal}{\emph{Ieee Access}}  \bibinfo{volume}{9} (\bibinfo{year}{2021}), \bibinfo{pages}{99495--99503}.
\newblock


\bibitem[Y.~Lee(2022)]%
        {ylee2022a}
\bibfield{author}{\bibinfo{person}{T.~You Y.~Lee, Y.~Kim}.} \bibinfo{year}{2022}\natexlab{}.
\newblock \showarticletitle{Analysis of QoQ GDP Prediction Performance Using Deep Learning Time Series Model}.
\newblock \bibinfo{journal}{\emph{KIISE}}  \bibinfo{volume}{49} (\bibinfo{year}{2022}), \bibinfo{pages}{873--883}.
\newblock


\bibitem[Yi et~al\mbox{.}(2022)]%
        {lee2022d}
\bibfield{author}{\bibinfo{person}{Hyun~Chang Yi}, \bibinfo{person}{Dongkyu Choi}, {and} \bibinfo{person}{Yonggun Kim}.} \bibinfo{year}{2022}\natexlab{}.
\newblock \showarticletitle{Dynamic Factor Model and Deep Learning Algorithm for GDP Nowcasting}.
\newblock \bibinfo{journal}{\emph{Bank of Korea Economic Analysis}}  \bibinfo{volume}{28} (\bibinfo{year}{2022}).
\newblock


\end{thebibliography}
\end{multicols}

\end{document}